\documentclass[letterpaper, 10pt]{ieeeconf_custom} 

\IEEEoverridecommandlockouts
\usepackage{cite}
\usepackage{amsmath,amssymb,amsfonts}
\usepackage{algorithmic}
\usepackage{graphicx}
\usepackage{textcomp}
\usepackage{xcolor, colortbl}
\usepackage{comment} 
\usepackage{pdfcomment} 
\usepackage{xspace}
\usepackage{caption} 
\usepackage{booktabs} 
\usepackage{float}
\usepackage{tikz}
\usepackage{wrapfig}
\usepackage{tikz}  
\usepackage{tikzpagenodes}  

\newcommand*\circled[1]{\tikz[baseline=(char.base)]{
            \node[shape=circle,draw,inner sep=1pt] (char) {#1};}}
          
\newcommand{\eg}{\emph{e.g.},\xspace}

\newcommand{\transp}{\mathrm{\sf T}}
\newcommand*{\inv}{^{-1}}

\newcommand{\cb}{B}
\newcommand{\ci}{I}
\newcommand{\cc}{C}
\newcommand{\cw}{W}
\newcommand{\Rw}{r_w}

\begin{document}
 \makeatletter
\newcommand{\linebreakand}{%
  \end{@IEEEauthorhalign}
  \hfill\mbox{}\par
  \mbox{}\hfill\begin{@IEEEauthorhalign}
}
\makeatother



\def\BibTeX{{\rm B\kern-.05em{\sc i\kern-.025em b}\kern-.08em
    T\kern-.1667em\lower.7ex\hbox{E}\kern-.125emX}}
   
\arrayrulecolor{darkgray}
\title{\LARGE \bf The Wheelbot: A Jumping Reaction Wheel Unicycle}

\author{%
\thanks{*This work was supported in part by the Max Planck Society and the Cyber Valley initiative. A. Ren\'e Geist was supported by the \href{https://imprs.is.mpg.de/}{IMPRS-IS}.}
A.~Ren\'e Geist$^{1,2}$, Jonathan Fiene$^{3}$, Naomi Tashiro$^{3}$, Zheng Jia$^{2}$, and Sebastian Trimpe$^{1,2}$
\thanks{$^{1}$Authors are with the Institute for Data Science in Mechanical Engineering, RWTH Aachen University, Germany, {\tt\footnotesize rene.geist@dsme.rwth-aachen.de} (Corresponding author), {\tt\footnotesize trimpe@dsme.rwth-aachen.de}}%
\thanks{$^{2}$Authors were with the Intelligent Control Systems Group,
Max Planck Institute for Intelligent Systems (MPI-IS),
Stuttgart, Germany}
\thanks{$^{3}$Authors are with the Robotics Central Scientific Facility,
MPI-IS, Stuttgart, Germany,
{\tt\footnotesize fiene@is.mpg.de}, {\tt\footnotesize tashiro@is.mpg.de}}
}

\maketitle

\begin{abstract}
Combining off-the-shelf components with 3D-printing, the Wheelbot is a symmetric reaction wheel unicycle that can jump onto its wheels from any initial position. With non-holonomic and under-actuated dynamics, as well as two coupled unstable degrees of freedom, the Wheelbot provides a challenging platform for nonlinear and data-driven control research. This paper presents the Wheelbot's mechanical and electrical design, its estimation and control algorithms, as well as  experiments\footnote[4]{A video of the Wheelbot and additional project details are provided at \href{https://sites.google.com/view/wheelbot}{https://sites.google.com/view/wheelbot}} demonstrating both self-erection and disturbance rejection while balancing.
\newline
\end{abstract}


    \begin{tikzpicture}[remember picture,overlay,shift={(current page text area.south west)}]
    \node [below right] {\parbox{\textwidth}{{\color{white}.}\\ \footnotesize \textbf{Accepted final version.}  To appear in \textit{IEEE Robotics and Automation Letters}.  \\ \textcopyright 2022 IEEE.  Personal use of this material is permitted.  Permission from IEEE must be obtained for all other uses, in any current or future media, including reprinting/republishing this material for advertising or promotional purposes, creating new collective works, for resale or redistribution to servers or lists, or reuse of any copyrighted component of this work in other works. \hfill}};
    \end{tikzpicture}

\section{Introduction}
Recent advancements in actuators, sensors, and embedded controllers, together with the explosion of low-cost 3D rapid prototyping, have enabled the development of a wide variety of novel robotic testbeds.  Such systems enrich the controls and robotics community by exposing unique compositions of dynamical properties, while also providing a platform for the exploration of novel mechatronic solutions.

When contemplating the architecture of a robotic testbed, its designer first encounters several system-level design decisions, including the number of degrees of freedom (DOF), as well as the layout of the mechanical and electrical components.  While a variety of interesting arrangements exist in such a large design space, this work focuses on the realm of nonlinearly-coupled, under-actuated systems, wherein the number of actuators is less than the DOF. A textbook example of such a system is the ``cart-pole pendulum'', which has only two DOF (the position and pendulum angle) and one actuator controlling the cart's position. The cart-pole pendulum's low-dimensional dynamics eases its analysis while its simple design reduces cost and maintenance. Yet, for systems with more DOF, numerous questions arise on how to identify and leverage coupling terms for control. In what follows, we propose a control testbed that offers challenging under-actuated dynamics with interesting dynamical properties while being compact in its design and resorting to off-the-shelf components.

\begin{figure}[t]
 \vspace{0.05cm}
     \centering
            \includegraphics[width=0.49\textwidth, trim={3.0cm 1.1cm 4.0cm 0.8cm},clip]{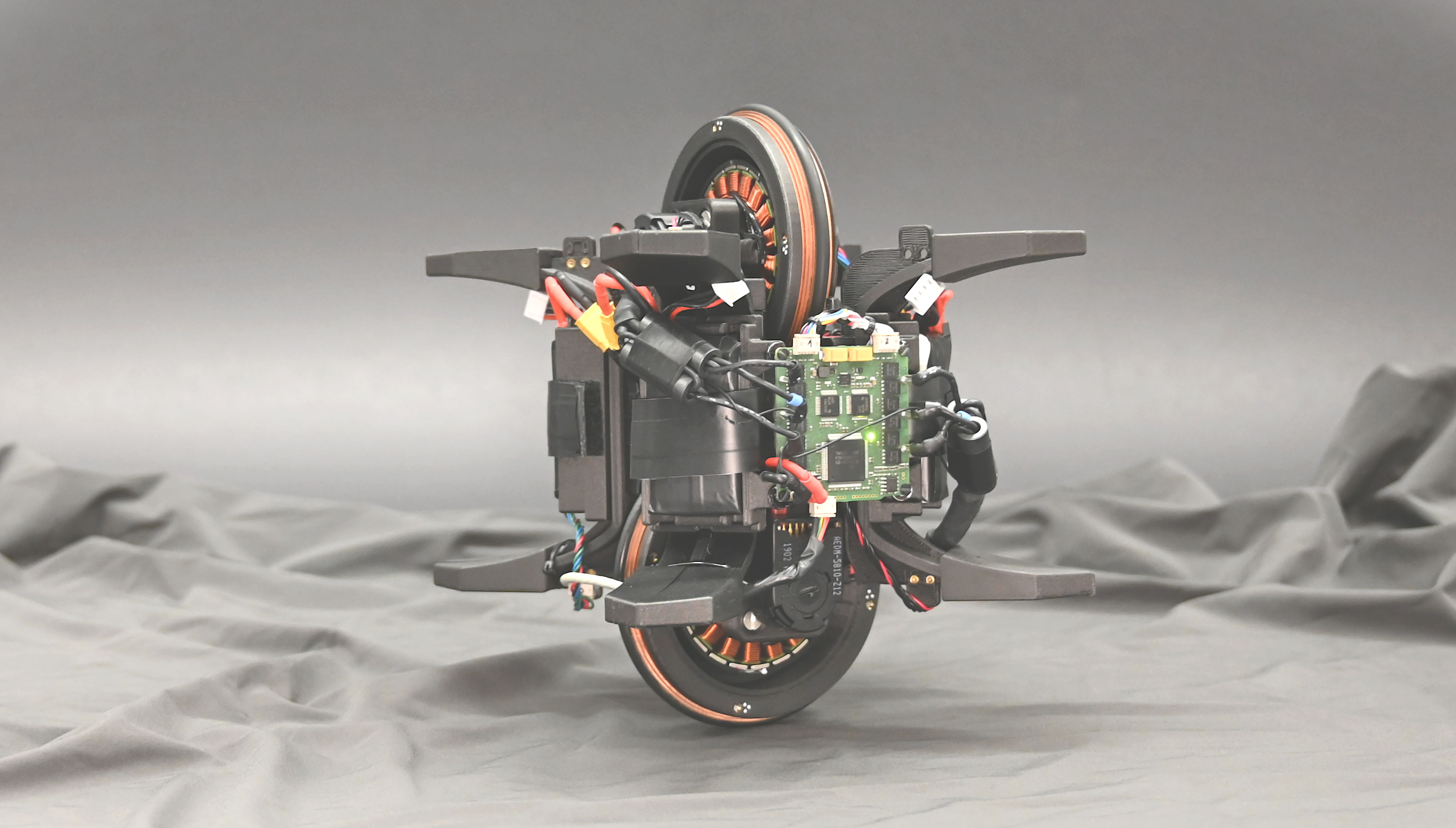}
 \caption{The Wheelbot is a reaction wheel-driven unicycle robot that uses brushless motors to self-erect after toppling.}
 \label{fig:wheelbot_intro}
\vspace{-0.3cm}
\end{figure}

\subsection{Self-balancing Wheeled Testbeds}
Typically utilizing off-the-shelf electric motors, wheeled robots are interesting control testbeds that usually require a small operating space while also providing a wide range of different dynamical properties.
The wheels of such robots can be distinguished into rolling wheels and reaction wheels. Rolling wheels leverage friction forces for locomotion and are found on Ballbots \cite{frankhauser2010ballbot, nagarajan2014ballbot} and Segway-esque robots such as the Ascento \cite{klemm2019ascento}. Reaction wheels apply free torques and are used in wheeled pendulums such as \cite{belascuen2018design} and cube-like structures such as \cite{gajamohan2013cubli}. 
From a dynamics perspective, balancing with a reaction wheel requires fast changes in the motor's direction of rotation and may also evoke high motor velocities. In turn, actuating a reaction wheel with an electric motor summons rate-dependent control constraints which may motivate further research on optimization-based control \cite{xiong2021slip}. 
In comparison, rolling wheels rotate at comparably lower speeds and may introduce non-holonomic kinematic constraints as in the case of the Ascento \cite{klemm2019ascento}. Moreover, using rolling wheels may require the linearized closed-loop system dynamics to be non-minimum phase as in the case of \cite{frankhauser2010ballbot, nagarajan2014ballbot, klemm2019ascento}. 

Brushless electric motors have enabled wheeled robots such as the Cubli \cite{gajamohan2013cubli, muehlebach2017nonlinear} and Ascento \cite{klemm2019ascento} to perform fast-changing maneuvers that are subject to discontinuous contact dynamics.  As such dynamics are difficult to model a-priori, linear control algorithms tend to perform below expectations, motivating the usage of different control strategies that either identify modeling errors from data or incorporate probabilistic nonlinear error functions into the control design.

\subsection{Designing an agile reaction wheel unicycle} \label{subsec:concept}
The Wheelbot arose from the desire for a compact, under-actuated control testbed with non-holonomic and fast-changing discontinuous dynamics that could be operated in a relatively small space. To meet these objectives, the small unicycle robot shown in Fig.\ \ref{fig:wheelbot_intro} was designed to include two actuated wheels attached to a rigid body, with one rolling on the ground and the other acting as a reaction wheel.

Previous literature proposes reaction wheel unicycle robots
whose rotation axis is either coaxial to a line connecting the center of both wheels  \cite{vos1992nonlinear,deisenroth2010efficient, rizal2015point} or orthogonal to such a line  \cite{lee2013unicyclecontroller, rosyidi2016speed, rizal2015point}. Unicycle robots with coaxially-oriented reaction wheels do not directly actuate the unstable roll DOF but instead turn the robot in the tilting direction while the rolling wheel prevents toppling. In contrast, an orthogonal-configuration unicycle directly actuates the roll DOF. 

When designing the Wheelbot, we opted for an orthogonal configuration to provide direct control of the unstable roll and pitch DOF. As the roll, pitch, and yaw dynamics are decoupled when linearized around the upright equilibrium, an orthogonal configuration allows one to tune the roll and pitch balancing controllers independently from each other. These controllers then act as a starting point for tackling more challenging research questions that revolve around the identification and control of the robot's yaw dynamics or, as a subsequent step, the control of the robot along a desired trajectory. 

Recently proposed designs for orthogonal-configuration unicycles \cite{lee2013unicyclecontroller, rosyidi2016speed, rizal2015point, 9483037} resort to large, high-inertia reaction wheels to reduce wheel acceleration and in turn avoid exposing the system's electronics to larger currents at faster wheel speeds. Yet, to be able to perform agile maneuvers and self-erect from any initial pose, the reaction wheel of the Wheelbot must be considerably smaller than those of the aforementioned unicycle robot designs, which led to several so far unvanquished design challenges.

To the best of our knowledge, the Wheelbot forms the first unicycle robot that can self-erect from any initial position, can recover from significant disturbances, is symmetric such that both wheels can act as reaction wheel or rolling wheel, and carries its own onboard power supply.

\section{Conceptual design}
The detailed design of the Wheelbot began with a number of high-level architectural decisions:

\begin{enumerate}
    \item The outer geometry of the robot should allow it to self-erect from any initial position.
    \item The robot's bottom and top half should be identically constructed, thereby reducing the number and complexity of the 3D-printed parts.
    \item The robot should carry its own power supply.
    \item Control of the robot should be shared between an on-board embedded processor and a wirelessly-connected supervisory controller.
\end{enumerate}
The realization of these points required solving numerous mechanical, electrical, and programmatic challenges. 

\begin{table}[t]
	\begin{minipage}{0.5\linewidth}
	\caption{System overview.}
		\label{tab:le}
		\centering
		\resizebox{\textwidth}{!}{%
\begin{tabular}{@{}l|l@{}}
\toprule
\textbf{Category} & \textbf{Value} \\ \midrule
Total weight & 1.4 kg \\ 
Wheel weight & 0.32 kg \\ 
Operating time & $\sim$20 min \\ 
Max. supply voltage & 25.2 V \\ 
Battery capacity & 1.3 Ah \\ 
Nominal motor current & 19 A \\ 
Nominal motor torque & 1.3 Nm \\ 
\bottomrule
\end{tabular}
}
	\end{minipage}
	\hspace{0.2cm}
	\begin{minipage}{0.46\linewidth}
	\vspace{2mm}
		\centering
		\includegraphics[width=0.98\textwidth]{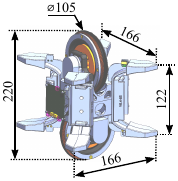}
		\captionof{figure}{Dimensions (mm).}
		\label{fig:dimensions}
	\end{minipage}
\end{table}

\begin{figure}[t]
    \centering
    \vspace{-1mm}
    \includegraphics[scale=0.75, trim = {0cm, 0cm, 0cm, 0cm}]{ 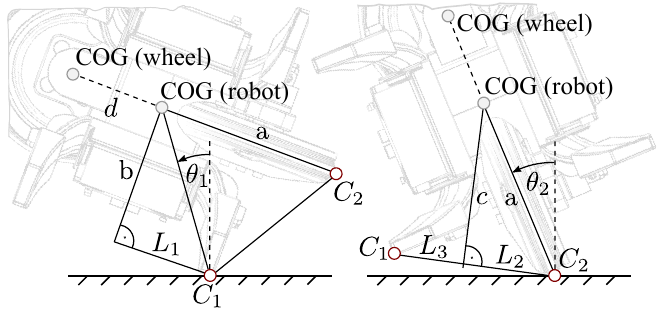}
    \vspace{-0.2mm}
    \caption{During the two stand-up steps, the system is modelled as a 2D reaction wheel pendulum rotating its centre of gravity (COG) either around $C_1$ changing $\theta_1$ or, $C_2$ changing $\theta_2$.}
    \label{fig:concept2}
\end{figure} 

\subsection{Modeling stand-up dynamics}
The ability to self-erect while carrying its own power supply requires the motor in the reaction wheel to apply considerable torques. Therefore, we first discuss the system's dynamics during self-erection maneuvers as these have been decisive for determining the required maximum motor torque and subsequently on choosing the robot's power supply.

To self-erect, the robot first accelerates one wheel until the rolling wheel has ground contact. Afterward, the robot rotates to its upright equilibrium position either using its reaction wheel, which we refer to as \emph{stand-up}, or its rolling wheel, which we refer to as \emph{roll-up}. The robot's dimensions are chosen to minimize the required motor torque during stand-up, as these torques are considerably larger than the motor torques required for a roll-up.
As illustrated in Figure \ref{fig:concept2}, the robot's center assembly is cube-shaped such that during a stand-up the system self-erects by rotating first around the point $C_1$ and then rotating around the point $C_2$. If during the stand-up these contact points slide over the ground, the necessary torque for self-erection reduces compared to the contact points remaining static. Therefore, the dynamics of the stand-up steps that would require the maximum motor torques are equivalent to the dynamics of a 2D inverse reaction wheel pendulum, cf.\, \cite{belascuen2018design}, writing
\begin{equation}
    \frac{d}{dt} \begin{bmatrix} 
    \dot \theta_i \\ \omega \end{bmatrix} = 
    \begin{bmatrix} 
    \left(Q_{g}(\theta_i) - Q_{\text{w}}(\omega)\right) / I_{\text{total},i} \\ Q_{\text{w}}(\omega) / I_{\text{w}}\end{bmatrix}, \label{eq:standup_ode}
\end{equation}
with the wheel rate $\omega$, the gravitational torque $Q_{g}$, motor torque $Q_{\text{w}}$, the wheel's rotational inertia around the rotation axis $I_{\text{w}}$, and the system's rotational inertia $I_{\text{total},i}$ with respect to the contact point $C_i$. 
Due to the cube shape, the reaction wheels can be decelerated before the second stand-up step by a maximum motor torque of $T_3 = L_3 m_{\text{total}} g$ with the system's total weight $m_{\text{total}}$ and gravitational acceleration $g$.

\subsection{Choosing the height to determine wheel inertia}
The design of the robot started with choosing its height. On the one hand, the system should be small to allow driving maneuvers in a reasonably sized laboratory while reducing the need for extensive safety precautions. On the other hand, if the robot is too small, its assembly becomes laborious. We chose the robot's height as $220\,\text{mm}=2a$.  After repeated iterations through the below detailed design procedure, the final wheel weight was set to 0.32\,kg. According to \eqref{eq:standup_ode}, $I_{\text{w}}$ determines how fast the motor torque reduces due to an increase in $\omega$. Therefore, the wheel design should maximize $I_{\text{w}}$ by placing its mass as far as possible from its rotation axis, effectively forming a  hollow cylinder with inertia \mbox{$I_{\text{w}} \approx 0.5 m_{\text{w}} \left(( \Rw - 8\,\text{mm})^2 + \Rw^2\right)$} and radius $\Rw = a - 4\,\text{mm}$.

\subsection{Determining the chassis geometry}
The required torque for the stand-up is significantly affected by the length $b$, $L_1$, and $L_2$. The length \mbox{$b=83\,\text{mm}$} provides just enough space to shelter the batteries, microcontroller, and cables inside the cube-shaped chassis. 
We then chose $L_1=L_2\approx 61\,\text{mm}$ to minimize the required motor torque that keeps the system in a static equilibrium \mbox{$|Q_{\text{w,LB}}|= \text{max}(L_1 m_{\text{total}} g, L_2 m_{\text{total}} g)=0.83\,\text{Nm}$}. Notably, $Q_{\text{w,LB}}$ forms a lower bound on the required motor torque during a stand-up.

\subsection{Choosing motor controller and motors}
 For a suitable brushless motor to apply a torque larger than $Q_{\text{w,LB}}$, a motor controller must be chosen that can sustain the required current while also being small enough to fit onto the robot. After a comparison of brushless motor controllers, we chose the \href{https://github.com/open-dynamic-robot-initiative/open_robot_actuator_hardware/blob/master/electronics/micro_driver_electronics/README.md#micro-driver-electronics}{``$\mu$Driver v.2''} from the open robotics initiative project \cite{grimminger2020open}, a compact dual-channel brushless motor controller based on the TI TMS320F28069 microcontroller and a pair of TI DRV8305 smart three-phase gate drivers. The $\mu$Driver v.2 can operate up to a maximum of 44\,V while delivering up to 30\,A per channel.  Its diagonal size of 70\,mm is considerably smaller than comparable alternatives such as the \href{https://odriverobotics.com/shop/odrive-v36}{ODrive v3.6} with a diagonal size of 150\,mm.

Given the system's geometry, estimates for the components' weights, and the specs of the motor controller, we compared commercially available brushless motors. Ideally, the reaction wheel motor should:
\begin{itemize}
    \item be light to reduce the required torque for stand-up,
    \item be slim to act as a hub motor for the wheels,
    \item provide larger stall torques for the stand-up at preferably lower currents and lower cost. 
\end{itemize}
We finally chose the T-motor Antigravity 6007 Kv160 brushless motor with a continuous torque of 1.3\,Nm at 18\,A. At a cost of 120\,€, this motor has a specific torque of $1.3 \text{Nm} / 0.18\,\text{kg}=7.2 \text{Nm} / \text{kg}$ at 18\,A, and a specific torque to cost ratio of $7.2 \frac{\text{Nm}}{kg} / 120\,\text{€}=0.06 \text{Nm} / \text{kg€}$.

\begin{figure}
      \centering
    \includegraphics[scale=0.65, trim = {0cm, 0cm, 0cm, 0.2cm}]{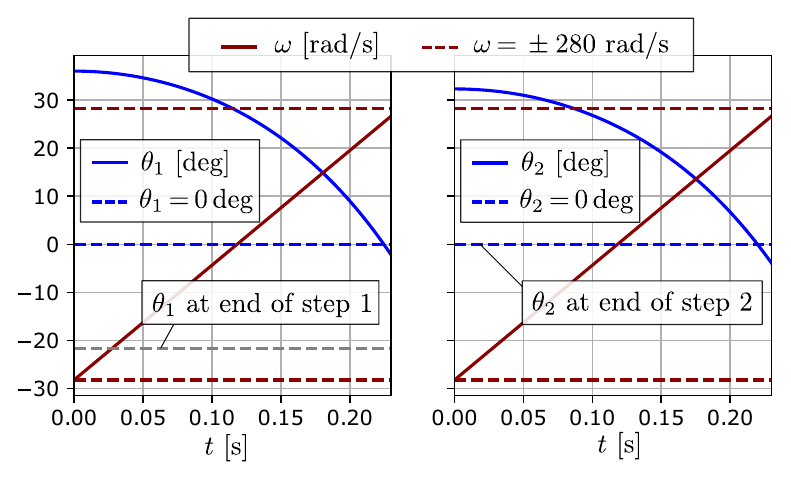}
    \caption{In the 2D reaction wheel pendulum model of the stand-up with an initial wheel speed of $\omega=-280\,\text{rad/s}$, the motor must apply a constant motor torque of $Q_{\text{W}}{=}1.2\,\text{Nm}$ to self-erect in two steps.}
    \label{fig:standup_test}
\end{figure} 

\subsection{Determining the supply voltage}
The chosen motor is rated to be operated using 6S to 12S LiPo batteries -- that is, at a voltage of around 22\,V to 44\,V. We opted for operating the motors at 22\,V to reduce the battery weight. Albeit, choosing a lower voltage for operating the motors comes at the price of the motor-controller only sustaining a current of 18\,A and a motor torque of 1.3\,Nm until a rotational motor velocity of $\omega {=} 282\,\text{rad/s}$ (2700 RPM) is reached. Subsequently, we tested if the chosen system is able to self-erect by numerically integrating the stand-up dynamics \eqref{eq:standup_ode} using the Wheelbot's geometry as depicted in Fig.\,\ref{fig:concept2}. The initial wheel velocity at the beginning of each step has been set to $-280\,\mathrm{rad/s}$ while assuming that the motor applies a constant torque of $Q_{\text{W}}=1.2\,\text{Nm}$. The corresponding simulation results as depicted in Figure \ref{fig:standup_test} show that the system's concept is able to self-erect. While in the first stand-up step the robot rotates by $58 \,\text{deg}$ compared to a rotation of $32 \,\text{deg}$ in the second step, the angle the system must  rotate until the gravitational torque helps in the first stand-up only amounts to $36 \,\text{deg}$.

\section{System Description}
Next, we describe how the core Wheelbot concept from the previous section is realized in terms of its mechanical, electronics, and software components.

\subsection{Mechanical design}
 The robot consists of two identical wheel assemblies mounted to a center frame. Most mechanical components were created from Onyx ABS plastic using a Markforged OnyxOne 3D printer. Threaded brass inserts are heat impressed into the 3D printed parts enabling the removal of screws without wear and tear.

\paragraph{Center frame}
The center assembly consists of a 3D-printed chassis sheltering microcontrollers, four batteries, cables, four inertia measurement units (IMUs), and the two wheel assemblies. The batteries are placed symmetrically with respect to the $x$ and $y$ axis of the body-fixed frame as they contribute considerably to the system's overall weight and inertia. The width and height of the chassis cube structure are provided in Fig.\ \ref{fig:dimensions}.

\begin{figure}[t]
     \centering
     \vspace{0.2cm}
       \includegraphics[width=0.35\textwidth]{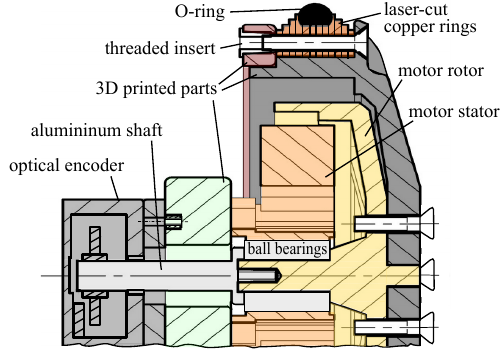}
 \caption{Sectional drawing of the wheel assembly.} \label{fig:overview_wheel_design}
 \vspace{-0.2cm}
\end{figure}

\paragraph{Wheel assembly}
As depicted in Fig.\,\ref{fig:overview_wheel_design}, each wheel assembly consists of several 3D-printed parts, a wheel, a brushless electric motor, and an optical encoder. 
For the wheels, we laser-cut 1 mm thick copper rings, which are then stacked onto a 3D printed hub attached to the wheel's motor as depicted in Fig.\,\ref{fig:overview_wheel_design} such that its rotational inertia amounts to $5\cdot10^{-4}$ kg-m$^2$.
A rubber O-ring (75~mm inside diameter, 5~mm cross-section diameter) was fitted to the wheel's outer rim to increase grip when touching the ground.

\subsection{Electronics design}
Table \ref{tab:le} gives a general system overview, while Table \ref{tab:component_table} provides a summary of the chosen electronic components.
Power is supplied by four 12.6\,V, 650\,mAh Lithium-Polymer (LiPo) batteries arranged as two parallel pairs of serially connected packs to provide a voltage of around 24\,V. 

 Motor commands and feedback signals are transmitted between the \href{https://github.com/open-dynamic-robot-initiative/open_robot_actuator_hardware/blob/master/electronics/micro_driver_electronics/README.md#micro-driver-electronics}{$\mu$Driver v.2} motor controller and a compact 16-MHz ATmega32U4-based MAEVARM M2 microcontroller via a high-speed Serial-Peripheral Interface (SPI) connection. 

The SPI interface of the M2 is also used for the collection of data from four TDK-Invensense ICM-20948 9-DOF IMUs attached to four corners of the cube structure. Each IMU provides triaxial acceleration, angular rate, and magnetic field measurements, though in the present configuration the magnetometer readings are not used due to high noise from the nearby brushless motors. To maximize sensitivity while avoiding saturation, the range of the accelerometers is set to $\pm 2g_0$ where $g_0$ denotes the gravitational acceleration constant $g_0\approx9.81\,\mathrm{m/s}^2$, while the gyroscopes are set to 500\,deg/s.
As shown in Fig.\,\ref{fig:overview_architecture}, the robot's state estimator and controllers are executed on the M2, which is also responsible for the receipt of user inputs via a Nordic nRF25LE1 2.4-GHz wireless link.

The use of brushless motors in such a compact design 
comes with the risk of significant Electro-Magentic Inference (EMI).  To reduce the emission of EMI, the motor cables are twisted and wrapped around ferrite rings to reduce high-frequency current oscillations. In addition, all SPI-communication cables are shielded with a grounded copper mesh and located as far from the motor cables as possible.

\subsection{Software design}
The embedded software for both the M2 and the motor driver is written in the ``C'' programming language. Running at a fixed loop rate of 100 Hz, the M2 receives user inputs over a wireless link, reads data from the four IMUs and two encoders, executes the state estimator and control routines, and outputs current values to the $\mu$Driver. 
The Wheelbot's user inputs and data are processed on a PC using the ``Python'' programming language via the ``Tkinter'' package. 

\section{Dynamics Modeling}
For the derivation of the state estimation and control algorithms, a dynamics model of the robot is derived from first principles. This dynamics model is also used for the system's simulation. The Lagrangian dynamics equations of a reaction wheel-driven unicycle robot are analogous to those of a pendulum-balanced unicycle as detailed in \cite{xu2017pendulumunicycle} if one replaces the pendulum's body with a reaction wheel. The reader is referred to \cite{xu2017pendulumunicycle} for a detailed description of the used coordinate systems.

\begin{table}[t]
\vspace{2mm}
 \caption{Component specifications.} \label{tab:component_table}
 \centering
\begin{tabular}{@{}l|l@{}}
\toprule
\textbf{Component} & \textbf{Specification} \\ \midrule
Motors & T-Motor Antigravity 6007 KV160 \\ 
Motor-controller & uDriver v2 \\ 
Additional $\mu$Controller & Maevarm M2 (ATmega32U4 processor) \\ 
IMUs & TDK-Invensense ICM-20948 9-DOF \\ 
Encoders & Avago Technologies AEDM5810Z12 \\ 
Batteries & LiPo - 11.1\,V (3S), 650\,mAh, 75\,C \\ \bottomrule
\end{tabular}
\end{table}

\begin{figure}[t]
     \centering
       \includegraphics[]{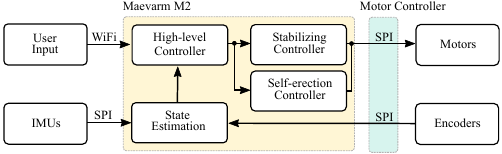}
 \caption{Overview of the software architecture.} \label{fig:overview_architecture}
\end{figure}

\begin{figure}[t]
\vspace{0.32cm}
     \centering
       \includegraphics[scale=0.95]{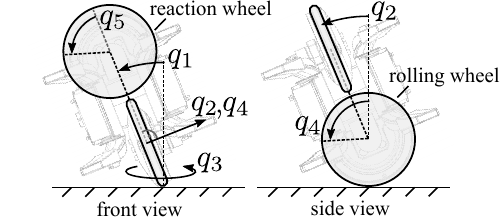}
 \caption{Generalized coordinates describing the system's pose.}
 \label{fig:freebody}
\end{figure}

\subsection{System coordinates}
The system's pose is described via the generalized coordinates $q \in \mathbb{R}^5$ with $q = [q_1, ... , q_5]^{\transp}$, the generalized velocity by $\dot{q}=\frac{dq}{dt}$, and the generalized acceleration as $\ddot{q}=\frac{d^2q}{dt^2}$. The body-fixed coordinate system is denoted as $\{ \cb \}$ with coordinate axes $\{e_1^{\cb}, e_2^{\cb}, e_3^{\cb}\}$ and origin $B$. As illustrated in Fig.\,\ref{fig:freebody} and Fig.\,\ref{fig:estiamtor-vectors}, the system coordinates are:
\begin{itemize}
    \item \emph{Contact point positions} $x$ and $y$: The Cartesian positions of $\{\cc\}$'s origin with respect to the inertial frame $\{\ci\}$.
    \item \emph{Roll angle} $q_1$: A rotation of $\{\cw\}$ around $e_1^{\cc}$.
    \item \emph{Pitch angle} $q_2$: A rotation of $\{\cb\}$ around $e_2^{\cw}$.
    \item \emph{Yaw angle} $q_3$: A rotation of $\{\cc\}$ around $e_3^{\ci}$.
    \item \emph{Rolling wheel angle} $q_4$: Describing the rolling wheel's rotation around $e_2^{\cw}$.
    \item \emph{Reaction wheel angle} $q_5$: Describing the reaction wheel's rotation around $e_1^{\cb}$.
\end{itemize}
 A vector is transformed from the inertial frame $\{\ci\}$ to the body-fixed frame $\{\cb\}$ by a yaw-roll-pitch Euler rotation sequence, reading
\begin{equation} \label{eq:inertia-to-body-transfo}
    R_{\cb\ci}(q_1,q_2,q_3) = R_2^\transp(q_2) R_1^\transp(q_1) R_3^\transp(q_3).
\end{equation}
The rotation from $\{\cb\}$ to $\{\ci\}$ is given as $R_{\ci\cb} = R_{\cb\ci}^\transp$. One obtains kinematic expressions for the system's position and orientation vectors using \eqref{eq:inertia-to-body-transfo}.

\subsection{Rigid-body dynamics and simulation model} \label{sec:simulation}
The derivation of the rigid-body dynamic equations is based on the following assumptions:
\begin{enumerate}
\item The system's bodies are rigid.
\item The rolling wheel rolls without slip.
\item Motors are not subject to friction or hysteresis effects.
\item Motor dynamics are neglected, being considerably faster than other dynamic terms.
\end{enumerate}
While some of these are clearly simplifying assumptions, the reality gap will be taken care of by feedback control, possibly combined with learning, which is one interesting challenge of this testbed.
With Assumption 2), the friction forces acting on the rolling wheel enforce an implicit non-holonomic constraint
    $[\dot{x} - \Rw \dot q_4  \cos(q_3),\: \dot{y} - \Rw \dot q_4  \sin(q_3)]^\transp = 0$, that reduces the generalized coordinates required to describe the system's motion to $q$.

As outlined in further detail in \cite{xu2017pendulumunicycle}, the system's equations of motion are obtained by solving the Euler-Lagrange equations. 
The dynamics equations were derived using Matlab's symbolic toolbox. These symbolic equations were then transferred to an s-function in Simulink.
In Simulink, the s-function dynamics model is combined with controllers as well as state-estimation routines. The controller model includes time delays in the obtained reference signal while the simulated IMU measurements are perturbed by Gaussian noise whose distribution closely resembles the characteristics of the real IMU noise. The Wheelbot's simulated motion is animated using a Simulink ``VR sink''.

\begin{figure}[t]
     \centering
     \vspace{0.15cm}
       \includegraphics[]{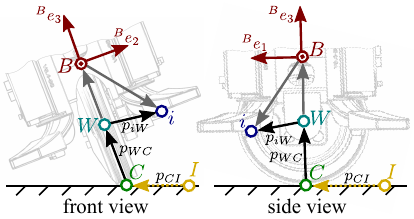}
 \caption{The locations of the robot's IMUs is described in terms of the coordinate systems $\{I\}$, $\{C\}$, $\{W\}$, and $\{B\}$.} \label{fig:estiamtor-vectors}
 \vspace{-0.2cm}
\end{figure}

\section{State Estimation}
An accurate and drift-free estimate of the robot's tilt angles $q_1$ and $q_2$ is critical for good control performance. In this section, we detail how the gyroscope and accelerometer-based tilt estimates are fused as illustrated in Fig.\ \ref{fig:filter_design}.

\subsection{Tilt and rate estimation using gyroscopes} \label{sec_gyroscope_tilt}
To obtain an estimate of the Euler rates $\{\dot q_1, \dot q_2, \dot q_3\}$, the measured gyro rates $^{i} \omega_i\in \mathbb R^3$ are first transformed into $\{\cb\}$, writing $^{\cb} \omega_i = R_{\cb i} {}^{i} \omega_i$, and afterwards using the previous tilt estimate $\{\hat q_1(k-1), \hat q_2(k-1)\}$ transformed into \{\ci\}, writing

\begin{equation} \label{eq:euler_rate_estimates}
\begin{bmatrix}
\dot{q}_{1, \mathrm{G}} \\
\dot{q}_{2, \mathrm{G}} \\
\dot{q}_{3, \mathrm{G}}
\end{bmatrix}=\begin{bmatrix}
R_2^{\transp} e_1 &
e_2 &
R_2^{\transp} R_1^{\transp} e_3
\end{bmatrix}^{-1} \sum_{i=1}^4 \frac{{ }^B \omega_i(k)}{4},
\end{equation}
where here, as well as in what follows, we use the abbreviation $R_1:=R_1(\hat q_1(k-1))$ and $R_2 :=  R_2(\hat q_2(k-1))$ and omit the discrete time step $k$.
Afterwards, the Euler rates in \eqref{eq:euler_rate_estimates} are integrated to obtain drifting pose estimates $\{q_{1,\text{G}},q_{2,\text{G}}, q_{3,\text{G}} \}$.

\subsection{Tilt estimation using accelerometers} \label{sec:accel_tilt}
A fundamental challenge in the computation of the tilt angle lies in distinguishing the gravitational acceleration from the acceleration that is caused by the system's motion. A common method for estimation of a robot's tilt angle are extended or unscented Kalman filters (KFs) \cite{state-estimation2013hertig, klemm2019ascento}. These approaches usually rely on local approximations of the attitude dynamics, which models the temporal correlation between the observed data. In turn, KFs approaches are susceptible to errors of the dynamics and noise parameters.

As an alternative to KFs, \cite{trimpe2010accelerometer} proposed an estimation algorithm for a rigid-body with a non-accelerated pivot point that has been used on various balancing robots \cite{trimpe2012balancing, gajamohan2013cubli, muehlebach2017nonlinear}. Given knowledge of the robot's kinematics model, the estimator in \cite{trimpe2010accelerometer} is computationally less demanding than KFs while providing a least-squares optimal estimation result. We extend the approach of \cite{trimpe2010accelerometer}, such that it may be used for wheeled balancing robots by estimating the pivot acceleration via encoder measurements.

An accelerometer measures an acceleration with respect to an observer in free fall. In turn, acceleration measurements of the $i$-th sensor at position $p_i$ with respect to $\{B\}$ read
\begin{equation} \label{eq:imu_meas}
    {}^{\cb}m_i = {}^{\cb} \ddot{p}_i - {}^{\cb}g + {}^{\cb}n_i, 
\end{equation}
with the acceleration due to the IMU's movement $\ddot{p}_i$ and measurement errors $n_i$.

\begin{figure}[t]
     \centering
    \includegraphics[width=8.8cm, trim={8.5cm 5.3cm 7cm 7cm}, clip]{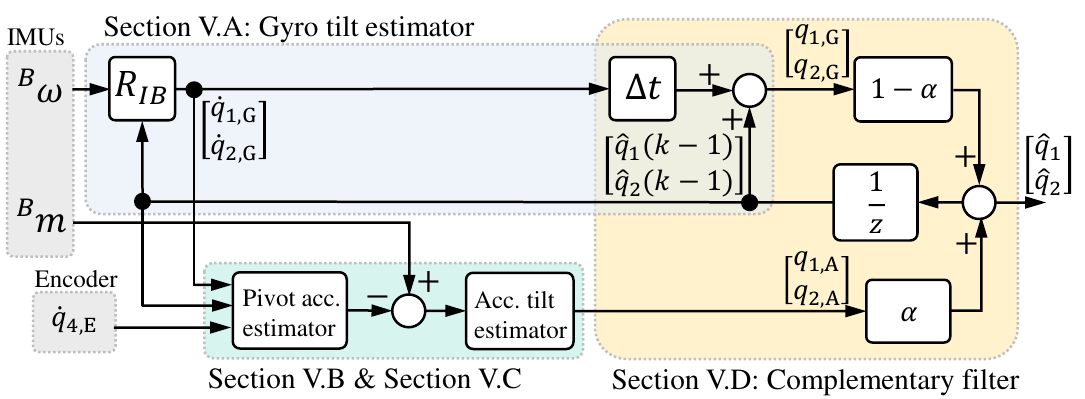}
 \caption{Block diagram of the tilt estimator.} \label{fig:filter_design}
\end{figure}

The approach in \cite{trimpe2010accelerometer} requires that the position vector $p_{i \cw}$ between the non-accelerated pivot point and each of the IMUs remains constant with respect to $\{\cb\}$. Consequently, we select the rolling wheel's center $W$ as the pivot point and estimate its acceleration $\ddot{p}_W$ as detailed in Section \ref{sec:pivot-acc}. By subtracting $\ddot{p}_W$ from ${}^{\cb}m_i$, we reduce \eqref{eq:imu_meas} to the problem of estimating the pose of a rigid-body with a non-accelerated pivot point, writing
\begin{equation} \label{eq:est1}
        {}^{\cb} \hat m_i \,\hat{=}\, {}^{\cb}m_i - {}^{\cb} \ddot{p}_W = {}^\cb \ddot p_{i \cw} - {}^{\cb}g + {}^{\cb}n_i, 
\end{equation}
with 
\begin{equation} \label{eq:est2}
    {}^\cb \ddot p_{i \cw} = {}^\cb \Omega {}^\cb p_{i \cw},
\end{equation}
where $p_{i \cw}$ points from the wheel center to the $i$-th IMU and ${}^\cb \ddot p_{i \cw} = R_{\cb \ci} \ddot R_{\ci \cb} {}^\cb p_{i \cw},$ with $\ddot R_{\ci \cb} = R_{\ci \cb} {}^\cb \Omega$. The matrix 
${}^\cb \Omega = {}^\cb \tilde \omega ^2+ {}^\cb \dot{ \tilde{\omega}}$,
represents the body's angular as well as centripetal acceleration with the angular velocity of $\{\cb\}$ relative to $\{\ci\}$ being $\omega$, and $\tilde \omega p_{i \cw} = \omega \times p_{i \cw}$.
By inserting \eqref{eq:est2} into \eqref{eq:est1}, one obtains a set of linear equations as
\begin{equation}
M=Q P+N,
\end{equation}
\begin{align}
M &=\left[\begin{array}{llll}
{}^\cb \hat m_1 & {}^\cb \hat m_2 & \ldots & {}^\cb \hat m_L
\end{array}\right] & \in \mathbb{R}^{3 \times L}, \\
Q &=\left[\begin{array}{llll}
{ }^{B} g & {}^\cb \Omega
\end{array}\right] & \in \mathbb{R}^{3 \times 4}\,, \\
P &=\left[\begin{array}{cccc}
1 & 1 & \ldots & 1 \\
{ }^{\cb} p_{1W} & { }^{\cb} p_{2W} & \ldots & { }^\cb p_{LW}
\end{array}\right] & \in \mathbb{R}^{4 \times L}, \\
N &=\left[\begin{array}{llll}
{ }^\cb n_{1} & { }^\cb n_{2} & \ldots & { }^\cb n_{L}
\end{array}\right] & \in \mathbb{R}^{3 \times L},
\end{align}
with the matrix of known sensor location $P$ having full row-rank by design. Then the optimal estimate for $Q$ is
\begin{equation} \label{eq:estimator}
    \hat{Q} = [{}^\cb \hat{g}, {}^\cb \hat{\Omega}] = M P^\transp (PP^\transp)\inv = M [X_1^\star, X_2^\star],
\end{equation}
with $X_1^\star \in \mathbb{R}^{L \times 1}$ being constant and only depending on the IMU positions. With \eqref{eq:inertia-to-body-transfo}, the gravitational acceleration reads
\begin{equation}
    {}^\cb g = R_2^\transp R_1^\transp {}^\ci g = g_0 
\begin{bmatrix}
-\cos (q_1) \sin (q_2)  \\
\sin (q_1) \\
\cos (q_1) \cos (q_2) 
 \end{bmatrix}.
\end{equation}
Finally, with ${}^\cb \hat{g}$ as in \eqref{eq:estimator}, the tilt angle estimates are
\begin{equation} \label{eq:final-tilt-estimate}
    q_{1, \text{A}} = \arctan \left( \frac{ {}^\cb \hat{g}_2 }{ \sqrt{{}^\cb \hat{g}_1^2 + {}^\cb \hat{g}_3^2}} \right),  q_{2, \text{A}} = \arctan \left( \frac{- {}^\cb \hat{g}_1}{ {}^\cb \hat{g}_3} \right).
\end{equation}

Muehlebach and D'Andrea \cite{muehlebach2017accelerometer} further extended \cite{trimpe2010accelerometer} by formulating \eqref{eq:imu_meas} using maximum likelihood estimation with the additional constraints that the  gravitational acceleration lies on a sphere of radius $g_0$, ${}^\cb \tilde \omega ^2$ is symmetric and ${}^\cb \dot{ \tilde{\omega}}$ is skew-symmetric. This leads to a non-convex optimization problem which can be solved iteratively. The estimator in \cite{muehlebach2017accelerometer} shows improved performance at the cost of a slightly higher computational load. The Wheelbot uses \cite{trimpe2010accelerometer} with our extension of estimating the pivot-point acceleration via encoder measurements as this estimator is sufficiently accurate.


\begin{figure}[t]
\vspace{1mm}
        \centering
    \includegraphics[width=0.45\textwidth, trim={0.2cm 0.0cm 0.0cm 0.0cm}]{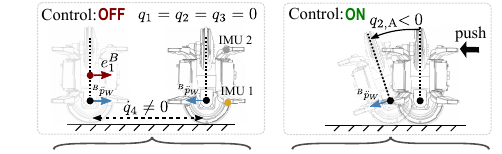}
    \includegraphics[width=0.47\textwidth, trim={0.7cm 0.5cm 1.6cm 0.0cm}]{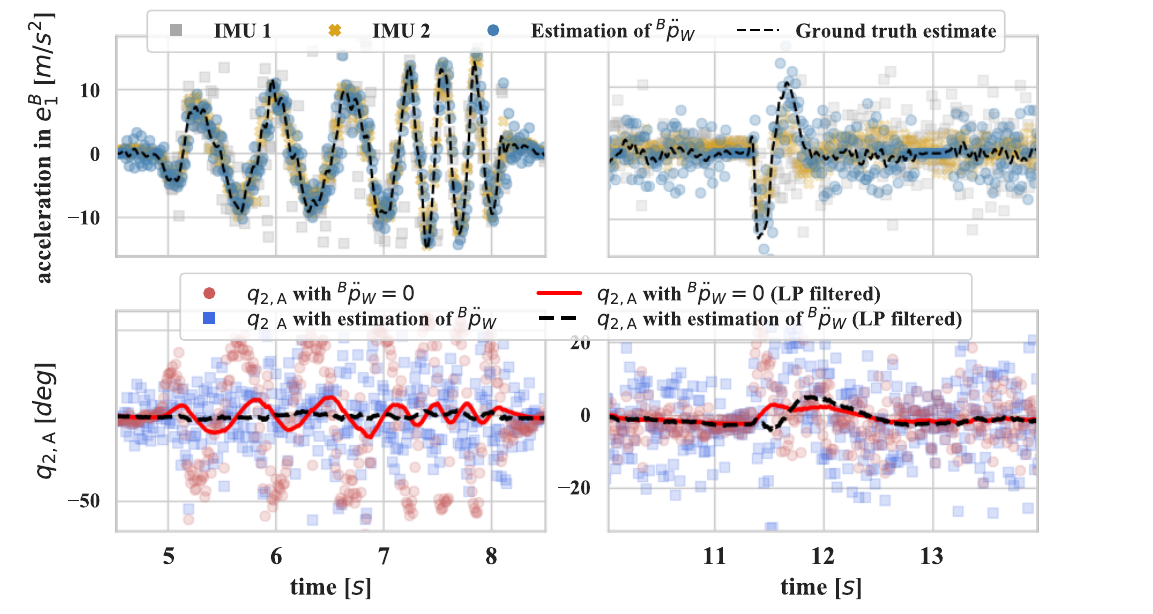} %
 \caption{Left:  The robot is translationally accelerated without rotating the system. Right: The balancing robot is pushed such that $q_2<0$. Top: Estimation of ${}^\cb \ddot p_{\cw}$ using $\ddot q_{4,\text{E}}$. Bottom: Pitch estimate $q_{2,\text{A}}$ with and without estimation of ${}^\cb \ddot p_{\cw}$. The lines are obtained by low pass filtering $q_{2,\text{A}}$.} \label{fig:pivot_estimation_data}
\end{figure}

\subsection{Pivot acceleration estimation using encoder} \label{sec:pivot-acc}
If we assume $\ddot{p}_W=0$ in the tilt angle estimation via \eqref{eq:estimator} and \eqref{eq:final-tilt-estimate}, a translational acceleration of the robot causes a non-zero tilt estimate even though the system did not tilt during the movement as shown in Fig.\,\ref{fig:pivot_estimation_data} (Left, red line).
On the Wheelbot, we reduce this error through estimation of $\ddot{p}_{\cw}$ and its insertion into \eqref{eq:est1}.  

To estimate $\ddot{p}_{\cw}$, we decompose the acceleration into its partial components as depicted in Fig.\,\ref{fig:estiamtor-vectors}, writing
\begin{equation} \label{eq:pw_sum}
    {}^\cb \ddot p_W = {}^\cb \ddot p_{\cc \ci} + {}^\cb \ddot p_{\cw \cc},
\end{equation}
where the vector $p_{\cc \ci}$ points from $I$ to the wheel contact point $C$, and $p_{\cw \cc}$ points from $C$ to $W$. Without slip, the acceleration at the ground contact point reads
\begin{equation}\label{eq:p_ci}
    {}^\cb \ddot p_{\cc \ci} =  R_2^\transp R_1^\transp \begin{bmatrix} \Rw \ddot q_4 & r_w \dot q_3 \dot q_4 & 0 \end{bmatrix}^{\transp}.
\end{equation}
Note that the encoder measures $\dot q_{4,\text{E}} = \dot q_4 - \dot q_2$. As $\dot q_4 \gg \dot q_2$, we assumed $\dot q_{4,\text{E}} \approx \dot q_4$. An estimate for $\ddot q_{4,\text{E}}$ is obtained by low-pass filtering $\dot q_{4,\text{E}}$ and resorting to numerical differentiation. 
The acceleration due to the change of $p_{\cw \cc}$ reads
\begin{equation} \label{eq_pwc}
    {}^\cb \ddot p_{\cw \cc} = R_2^\transp R_1^\transp  \begin{bmatrix}
    2 r_w\cos(q_1) \dot q_1 \dot q_3 + r_w \sin(q_1) \ddot q_3 \\
    r_w \sin(q_1) (\dot q_1^2 + \dot q_3^2) -r_w \cos(q_1) \ddot q_1 \\
    -r_w \cos(q_1) \dot q_1^2 - r_w \sin(q_1) \ddot q_1
    \end{bmatrix}.
\end{equation}

To determine the significance of the individual acceleration terms in \eqref{eq:pw_sum}, we ran a simulation of the Wheelbot in which the controller has been periodically excited resulting in a pirouette-esque motion. This simulation as well as further experiments on the real robot indicate that in \eqref{eq:pw_sum} only the term depending on $\ddot q_4$ significantly contributes to $\ddot p_w$ such that the other terms are being omitted for the estimation of the pivot acceleration.

Figure \ref{fig:pivot_estimation_data} illustrates the influence of the pivot acceleration estimation on the computation of the pitch angle. The ground truth estimate of ${}^\cb \ddot p_W$ is obtained by applying a non-causal low-pass filter with a cutoff frequency of 60\,Hz to $r_w\ddot q_{4,\text{E}}$. The first-order low-pass used to filter $q_{2,\mathrm{A}}$ has a cutoff frequency of 0.32\,Hz. This cutoff frequency matches the cutoff frequency of the low-pass filter part of the complementary filter that is used to filter the $q_{2,\mathrm{A}}$ estimates. 
In Fig.\, \ref{fig:pivot_estimation_data} (Left), the robot is held at $q_1=q_2=q_3=0$ while being only translational accelerated. In turn, the IMUs directly measure ${}^\cb \ddot p_W=e_1\Rw \ddot q_4$. 
Notably, the low-pass filtered estimate of $q_{2,\mathrm{A}}$ with estimation of ${}^\cb \ddot p_w$ remains close to zero.
In Fig.\,\ref{fig:pivot_estimation_data} (Right), the robot is pushed such that $q_2$ turns negative and then positive while the controller overshoots. Notably, the low-pass filtered $q_{2,\mathrm{A}}$ with estimation of ${}^\cb \ddot p_W$ reproduces the correct qualitative shape of $q_2$ during the push. 

\subsection{Sensor fusion via complementary filtering} \label{sec:sensor_fusion}
The gyroscope's tilt estimate $\{q_{1,\text{G}},q_{2,\text{G}}\}$ as in Section \ref{sec_gyroscope_tilt} is combined with the accelerometer's tilt estimate $\{q_{1,\text{A}},q_{2,\text{A}}\}$, cf.\ Section \ref{sec:accel_tilt} and  \ref{sec:pivot-acc}), via 
\begin{equation} \label{eq_complementary}
    \begin{bmatrix} \hat q_1 \\\hat q_2 \end{bmatrix} = \alpha \begin{bmatrix} q_{1,\text{A}} \\ q_{2,\text{A}} \end{bmatrix} + (1-\alpha) \begin{bmatrix} q_{1,\text{G}} \\ q_{2,\text{G}} \end{bmatrix},
\end{equation}
with fusion parameter $\alpha=0.02$. Equation \eqref{eq_complementary} corresponds to a discrete time complementary filter that combines the high-frequent part of the gyroscope tilt estimate with the low-frequent part of the accelerometer tilt estimate \cite{brown1997introduction}.

\section{Control} \label{sec:control} 
In what follows, we propose a baseline controller for the system as well as demonstrate its ability to jump onto one of its wheels and balance.

\subsection{Balancing Control}
 Linear quadratic regulator (LQR) control synthesis has been used successfully for the control of unstable wheeled robots \cite{lee2013unicyclecontroller, frankhauser2010ballbot, klemm2019ascento, li2013advanced-pendulum-control}. 
To obtain an LQR for balancing, the system's nonlinear dynamics are linearized around the upright equilibrium position ($q=\mathbf 0, \dot q = \mathbf 0$), to yield a linear state space model as
\begin{equation} \label{eq:linear_dynamics}
    \frac{d}{dt}  \begin{bmatrix}q_{I} \\ q_{II} \end{bmatrix} = \begin{bmatrix} A_{1} & \textbf{0}\\ \textbf{0} & A_{2} \end{bmatrix} \begin{bmatrix}q_{I} \\ q_{II} \end{bmatrix}  + \begin{bmatrix} B_1 && \mathbf 0 \\ \mathbf 0 && B_2 \end{bmatrix} \begin{bmatrix} u_1 \\ u_2 \end{bmatrix},
\end{equation}
with $q_{I}=[q_1, \dot q_1, q_5, \dot q_5]$, $q_{II}=[q_2, \dot q_2, q_4, \dot q_4]$, and system matrices of appropriate size. Due the linearization around the upright equilibrium, the roll and pitch dynamics in \eqref{eq:linear_dynamics} are decoupled. As in the linearized dynamics, the yaw angle $q_3$ and its velocity $\dot q_3$ are not controllable, the corresponding dynamic modes are being omitted from \eqref{eq:linear_dynamics}. 
Albeit, we observed in the experiments that some control actions lead to considerably agile movements along the yaw DOF. Additionally, the system's nonlinear dynamics model possesses over numerous coupling terms between the yaw dynamics and the other states, indicating that viewed from a nonlinear system's perspective the yaw dynamics can be controlled. However, an analysis of the system's nonlinear yaw dynamics exceeds the scope of this work.

The LQR's positive semi-definite weighting matrices for the roll controller and pitch controller are chosen as diagonal matrices. The diagonal elements of the $Q$ and $R$ matrices form tuning knobs that increase the quadratic cost of the respective state error or control input. 
With the $A$ and $B$ matrices from \eqref{eq:linear_dynamics}, carefully chosen $\{Q,R\}$ matrices, and the system's sample-time of $0.01 s$, the LQR gains for the discrete-time system are obtained as \eg
    $K_1 = [4.5,\, 0.25,\, 0.0003, \,0.0018]$ and $K_2 = [1.6,\, 0.14,\, 0.04,\, 0.0344]$,
such that
\begin{align} \label{eq:controller_final}
   u_1 = K_1 \begin{bmatrix} \hat q_1 - \bar q_1 & \dot q_{1,\text{G}} & q_{5, \text{E}} & \dot q_{5, \text{E}} \end{bmatrix}^{\transp}, \nonumber\\
      u_2 = K_2 \begin{bmatrix} \hat q_2 - \bar q_2 & \dot q_{2,\text{G}} & q_{4, \text{E}} & \dot q_{4, \text{E}} \end{bmatrix}^{\transp},
\end{align}
with the estimator tilt angle bias $\{\bar q_1 , \bar q_2 \}$ which has been estimated via a calibration routine before start.

\begin{figure}[t]
    \centering
    \vspace{0.05cm}
    \includegraphics[width=0.48\textwidth]{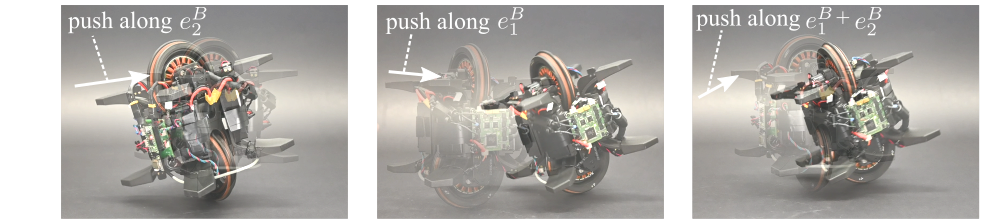}
    \includegraphics[width=0.48\textwidth, trim={0.0cm 0.0cm 0.0cm 0.0cm}]{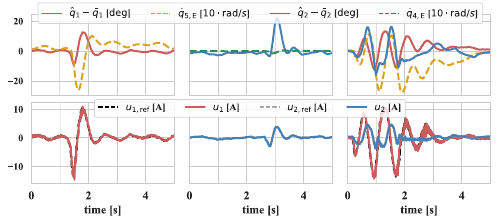}
 \caption{Balancing controller rejecting different pushes.} \label{fig:disturbance-rejection}
\end{figure}

Fig.\,\ref{fig:disturbance-rejection} illustrates how the Wheelbot rejects pushes coming from different directions with respect to $\{\cb\}$. 
With an estimated motor torque constant of $K_{\mathrm{T}}=0.075$, the motors applied $u \approx K_{\mathrm{T}} \cdot 13 \mathrm{A} = 1\,\mathrm{Nm}$ to reject the disturbances.

 \subsection{Jump-up Maneuvers} \label{sec:stand_up}
Several aspects aggravate the Wheelbot's self-erection maneuvers. Self-erection takes less than half a second at a control frequency of 100 Hz.
During self-erection, the robot's ground contact point changes altering its dynamics. When the wheel hits the ground, discontinuities and stick-slip effects may occur. 
While similar challenges are often encountered in other systems such as legged robots, the Wheelbot provides a compact and low-dimensional system for research on iterative learning and repetitive control algorithms.
Here, as soon as a balancing fails, the robot can self-erect and the experiment is continued.

\begin{figure}[t]
    \centering
    \vspace{0.05cm}
    \includegraphics[width=0.48\textwidth]{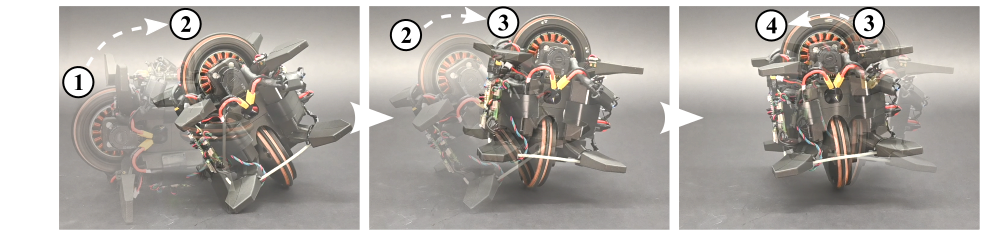}
    \includegraphics[trim={0.0cm 0.2cm 0.0cm 0.0cm}, width=0.48\textwidth]{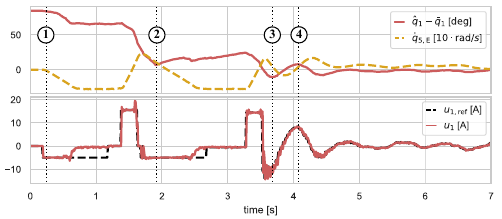}
 \caption{Stand-up experiment.} \label{fig:standup-experiment}
 \vspace{-0.1cm}
\end{figure}

\begin{figure}[t]
    \centering
    \includegraphics[width=0.48\textwidth]{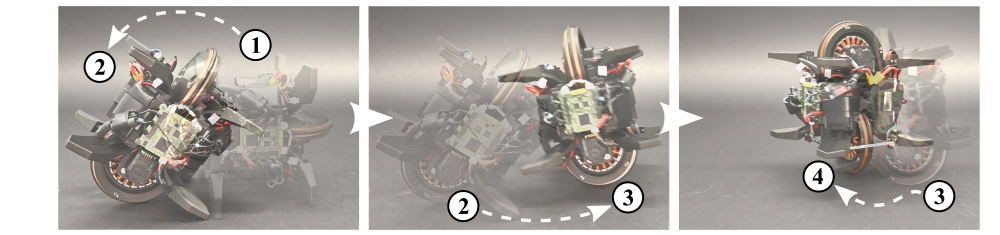}
    \includegraphics[trim={0.0cm 0.2cm 0.0cm 0.0cm}, width=0.48\textwidth]{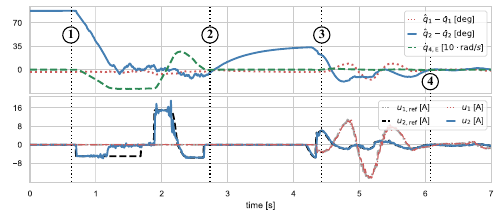}
 \caption{Roll-up experiment.} \label{fig:rollup-experiment}
 \vspace{-0.4cm}
\end{figure}

\subsubsection*{Stand-up}
Figure \ref{fig:standup-experiment} depicts the experimental results of a two-step stand-up. First, the reaction wheel accelerates until ${\hat{q}_1\,{\approx}\,30\,\mathrm{deg}}$ (Fig.\ \ref{fig:standup-experiment}, \circled{2}), then accelerates until ${\hat{q}_1\,{\approx}\,0\,\mathrm{deg}}$ (Fig.\ \ref{fig:standup-experiment}, \circled{3}) and finally switches to the roll balancing controller (Fig.\ \ref{fig:standup-experiment}, \circled{4}). 
At the beginning of the stand-up, the normal force between the ground and wheel does not suffice to prevent slip. Therefore, the pitch balancing controller is only switched on when the robot is nearly upright.

\subsubsection*{Roll-up}
As depicted in Fig.\ \ref{fig:rollup-experiment}, during a roll-up, the rolling-wheel first applies torques $u_2$ to rotate the robot such that it gets in contact with the ground. Secondly, the rolling wheel rotates the system to its upright equilibrium position. For the roll-up maneuver to be physically realizable, the normal force between the robot's wheel and the ground must be sufficiently large to prevent slip. The robot is able to perform a roll-up both on smooth acrylic glass as well as on a rubber mat.
The rolling wheel is quickly decelerated before establishing surface contact to prevent the wheel's O-ring from being damaged. 

\section{Conclusion}
This paper presents the design, modeling, and control of the Wheelbot, a novel reaction wheel unicycle robot. Besides the Wheelbot's ability to jump, its characteristic feature forms the symmetric design with two wheels that can act as reaction or rolling wheel depending on the configuration. The robot has been designed as a testbed for both continuous control (balancing, driving) and discrete control tasks (roll-up, stand-up), which we will explore in future research.
We hope that the Wheelbot's compact design and challenging composition of dynamical properties inspire research on nonlinear learning control as well as assist in the education of students interested in robotics and control systems.

\bibliographystyle{IEEEtran}
\bibliography{root.bib} 

\begin{thebibliography}{10}
\providecommand{\url}[1]{#1}
\csname url@rmstyle\endcsname
\providecommand{\newblock}{\relax}
\providecommand{\bibinfo}[2]{#2}
\providecommand\BIBentrySTDinterwordspacing{\spaceskip=0pt\relax}
\providecommand\BIBentryALTinterwordstretchfactor{4}
\providecommand\BIBentryALTinterwordspacing{\spaceskip=\fontdimen2\font plus
\BIBentryALTinterwordstretchfactor\fontdimen3\font minus
  \fontdimen4\font\relax}
\providecommand\BIBforeignlanguage[2]{{%
\expandafter\ifx\csname l@#1\endcsname\relax
\typeout{** WARNING: IEEEtran.bst: No hyphenation pattern has been}%
\typeout{** loaded for the language `#1'. Using the pattern for}%
\typeout{** the default language instead.}%
\else
\language=\csname l@#1\endcsname
\fi
#2}}

\bibitem{frankhauser2010ballbot}
P.~Fankhauser and C.~Gwerder, ``Modeling and control of a ballbot,'' ETH
  Z{\"u}rich, Tech. Rep., 06 2010.

\bibitem{nagarajan2014ballbot}
U.~Nagarajan, G.~Kantor, and R.~Hollis, ``The ballbot: An omnidirectional
  balancing mobile robot,'' \emph{The International Journal of Robotics
  Research}, vol.~33, no.~6, pp. 917--930, May 2014.

\bibitem{klemm2019ascento}
V.~Klemm, A.~Morra, C.~Salzmann, F.~Tschopp, K.~Bodie, L.~Gulich, N.~Kung,
  D.~Mannhart, C.~Pfister, M.~Vierneisel, F.~Weber, R.~Deuber, and R.~Siegwart,
  ``Ascento: A two-wheeled jumping robot,'' in \emph{International Conference
  on Robotics and Automation}.\hskip 1em plus 0.5em minus 0.4em\relax IEEE, May
  2019, pp. 7515--7521.

\bibitem{belascuen2018design}
G.~Belascuen and N.~Aguilar, ``Design, modeling and control of a reaction wheel
  balanced inverted pendulum,'' in \emph{2018 IEEE Biennial Congress of
  Argentina (ARGENCON)}.\hskip 1em plus 0.5em minus 0.4em\relax IEEE, 2018, pp.
  1--9.

\bibitem{gajamohan2013cubli}
M.~Gajamohan, M.~Muehlebach, T.~Widmer, and R.~D'Andrea, ``The cubli: A
  reaction wheel based 3d inverted pendulum,'' \emph{European Control
  Conference}, pp. 268--274, 2013.

\bibitem{xiong2021slip}
X.~Xiong and A.~Ames, ``Slip walking over rough terrain via h-lip stepping and
  backstepping-barrier function inspired quadratic program,'' \emph{IEEE
  Robotics and Automation Letters}, vol.~6, no.~2, pp. 2122--2129, 2021.

\bibitem{muehlebach2017nonlinear}
M.~Muehlebach and R.~D'Andrea, ``Nonlinear analysis and control of a
  reaction-wheel-based 3-d inverted pendulum,'' \emph{Transactions on Control
  Systems Technology}, vol.~25, no.~1, pp. 235--246, Jan 2017.

\bibitem{vos1992nonlinear}
D.~W. Vos, ``Nonlinear control of an autonomous unicycle robot: practical
  issues,'' Ph.D. dissertation, MIT, 1992.

\bibitem{deisenroth2010efficient}
M.~P. Deisenroth, ``Efficient reinforcement learning using gaussian
  processes,'' Ph.D. dissertation, 2010.

\bibitem{rizal2015point}
Y.~Rizal, C.-T. Ke, and M.-T. Ho, ``Point-to-point motion control of a unicycle
  robot: Design, implementation, and validation,'' in \emph{International
  Conference on Robotics and Automation}.\hskip 1em plus 0.5em minus
  0.4em\relax IEEE, May 2015, pp. 4379--4384.

\bibitem{lee2013unicyclecontroller}
J.~Lee, S.~Han, and J.~Lee, ``Decoupled dynamic control for pitch and roll axes
  of the unicycle robot,'' \emph{Transactions on Industrial Electronics},
  vol.~60, no.~9, pp. 3814--3822, Sep 2013.

\bibitem{rosyidi2016speed}
M.~A. Rosyidi, E.~H. Binugroho, S.~E.~R. Charel, R.~S. Dewanto, and
  D.~Pramadihanto, ``Speed and balancing control for unicycle robot,'' in
  \emph{International Electronics Symposium}.\hskip 1em plus 0.5em minus
  0.4em\relax IEEE, 2016, pp. 19--24.

\bibitem{9483037}
G.~P. Neves and B.~A. Angélico, ``A discrete lqr applied to a self-balancing
  reaction wheel unicycle: Modeling, construction and control,'' in \emph{2021
  American Control Conference}, 2021, pp. 777--782.

\bibitem{grimminger2020open}
F.~{Grimminger}, A.~{Meduri}, M.~{Khadiv}, J.~{Viereck}, M.~{Wüthrich},
  M.~{Naveau}, V.~{Berenz}, S.~{Heim}, F.~{Widmaier}, T.~{Flayols}, J.~{Fiene},
  A.~{Badri-Spröwitz}, and L.~{Righetti}, ``An open torque-controlled modular
  robot architecture for legged locomotion research,'' \emph{IEEE Robotics and
  Automation Letters}, vol.~5, no.~2, pp. 3650--3657, 2020.

\bibitem{xu2017pendulumunicycle}
Y.~Daud, A.~Al~Mamun, and J.-X. Xu, ``Dynamic modeling and characteristics
  analysis of lateral-pendulum unicycle robot,'' \emph{Robotica}, vol.~35,
  no.~3, pp. 537--568, Mar 2017.

\bibitem{state-estimation2013hertig}
L.~Hertig, D.~Schindler, M.~Bloesch, C.~D. Remy, and R.~Siegwart, ``Unified
  state estimation for a ballbot,'' in \emph{International Conference on
  Robotics and Automation}.\hskip 1em plus 0.5em minus 0.4em\relax IEEE, May
  2013, pp. 2471--2476.

\bibitem{trimpe2010accelerometer}
S.~Trimpe and R.~D'Andrea, ``Accelerometer-based tilt estimation of a rigid
  body with only rotational degrees of freedom,'' in \emph{International
  Conference on Robotics and Automation}.\hskip 1em plus 0.5em minus
  0.4em\relax IEEE, 2010, pp. 2630--2636.

\bibitem{trimpe2012balancing}
S.~Trimpe and R.~D’Andrea, ``The balancing cube: A dynamic sculpture as test
  bed for distributed estimation and control,'' \emph{IEEE Control Systems},
  vol.~32, no.~6, pp. 48--75, Dec 2012.

\bibitem{muehlebach2017accelerometer}
M.~Muehlebach and R.~D'Andrea, ``Accelerometer-based tilt determination for
  rigid bodies with a nonaccelerated pivot point,'' in \emph{Transactions on
  Control Systems Technology}, vol.~26, no.~6.\hskip 1em plus 0.5em minus
  0.4em\relax IEEE, Nov 2018, pp. 2106--2120.

\bibitem{brown1997introduction}
R.~G. Brown and P.~Y. Hwang, \emph{Introduction to random signals and applied
  Kalman filtering}.\hskip 1em plus 0.5em minus 0.4em\relax New York: John
  Wiley, 1997, vol. 3rd ed.

\bibitem{li2013advanced-pendulum-control}
Z.~Li, C.~Yang, and L.~Fan, \emph{Advanced Control of Wheeled Inverted Pendulum
  Systems}, 04 2014.

\end{thebibliography}

\end{document}